\documentclass{article}

\PassOptionsToPackage{numbers, compress}{natbib}

    \usepackage[preprint]{neurips_2025}

\usepackage[utf8]{inputenc} 
\usepackage[T1]{fontenc}    
\usepackage{hyperref}       
\usepackage{url}            
\usepackage{booktabs}       
\usepackage{amsfonts}       
\usepackage{nicefrac}       
\usepackage{microtype}      
\usepackage{xcolor}         

\usepackage{amsmath}
\usepackage[pdftex]{graphicx}

\title{ICEPool: Enhancing Graph Pooling Networks with Inter-cluster Connectivity}

\author{%
  Michael Yang \\
  Department of Computer Science\\
  University of California, Santa Barbara\\
  \texttt{yang335@ucsb.edu} \\
}

\begin{document}

\maketitle

\begin{abstract}
  Hierarchical Pooling Models have demonstrated strong performance in classifying graph-structured data. While numerous innovative methods have been proposed to design cluster assignments and coarsening strategies, the relationships between clusters are often overlooked. In this paper, we introduce Inter-cluster Connectivity Enhancement Pooling (ICEPool), a novel hierarchical pooling framework designed to enhance model's understanding of inter-cluster connectivity and ability of preserving the structural integrity in the original graph. ICEPool is compatible with a wide range of pooling-based GNN models. The deployment of ICEPool as an enhancement to existing models effectively combines the strengths of the original model with ICEPool's capability to emphasize the integration of inter-cluster connectivity, resulting in a more comprehensive and robust graph-level representation. Moreover, we make theoretical analysis to ICEPool's ability of graph reconstruction to demonstrate its effectiveness in learning inter-cluster relationship that is overlooked by conventional models. Finally, the experimental results show the compatibility of ICEPool with wide varieties of models and its potential to boost the performance of existing graph neural network architectures.
\end{abstract}

\section{Introduction}

Graphs are extensively utilized to represent connections and relationships among objects in the real world, with applications spanning social networks \cite{li2019semi}, academic networks \cite{sun2020pairwise, zhang2019oag}, anomaly detection \cite{akoglu2015graph, noble2003graph}, bioinformatics \cite{dobson2003distinguishing, toivonen2003statistical}, and more. The emergence of Graph Neural Networks (GNNs) has significantly enhanced models' ability to learn from such structured data, enabling state-of-the-art performance in tasks like node node classification \cite{hamilton2017inductive, kipf2016semi, zeng2019graphsaint}, link prediction \cite{zhang2018link}, and graph classification \cite{bruna2013spectral, defferrard2016convolutional, duvenaud2015convolutional, errica2019fair, xu2018powerful}. Recent advances have introduced attention mechanisms into graph learning, enabling nodes to selectively aggregate information \cite{velivckovic2017graph, wang2019heterogeneous}. In parallel, graph transformer architectures have emerged to effectively capture long-range dependencies and global structural patterns \cite{yun2019graph, wang2024graph}.

For graph-level tasks such as classification, Hierarchical Pooling Models have emerged as state-of-the-art approaches due to their ability to capture multi-scale structures within graphs. Traditional models like DiffPool \cite{ying2018hierarchical} and MinCutPool \cite{bianchi2020spectral} introduced effective clustering-based coarsening strategies, while more recent methods such as SEPG \cite{wu2022structural} and Cluster-GT \cite{huang2024cluster} further enhance flexibility in graph coarsening by incorporating adaptive mechanisms and structural priors.

However, a crucial limitation remains: existing Hierarchical Pooling Models often overlook the interconnection between clusters \cite{huang2024cluster}. In many real-world applications, such as understanding interactions between communities in social networks \cite{fan2019graph} or pathways in biological systems \cite{dobson2003distinguishing, morris2020tudataset}, the links between high-level graph regions carry essential information. Ignoring such connectivity can degrade the model’s ability to capture the true global structure and lead to loss of performance. In this work, we propose solutions to integrate the overlooked inter-cluster structural information. First, we integrated \textbf{Connection Entropy} into edge-conditioned GNN layers during the message propagation stage to capture connection patterns between clusters. Additionally, we redesign the aggregation process by introducing \textbf{SVDPool}, which captures the global structural patterns instead of restricting the aggregation to isolated clusters. The main contributions of this paper are summarized as follows:
\begin{itemize}
    \item To the best of our knowledge, this is the first work to improve hierarchical graph pooling by explicitly modeling and leveraging inter-cluster connectivity.
    \item We propose methods to re-evaluate the importance of nodes when aggregating the clusters' features and introduce edge features to capture the cluster-level connection patterns.
    \item We conduct extensive experiments on various datasets and models to demonstrate the performance boost of our methods as well as the superiority over baseline approaches.
\end{itemize}

\section{Related Work}
\label{related_work}

\subsection{Graph Pooling Layer}

Graph pooling operation is a critical component of Hierarchical Pooling Models, enabling the aggregation of node features and structural information into compact representations. Pooling methods are broadly categorized into node-based and cluster-based strategies:

\textbf{Node-based pooling methods}, such as Top-K pooling \cite{gao2019graph} and SAGPool \cite{lee2019self}, compute importance scores for nodes and select the top-ranking subset. These selected nodes are either directly used for downstream tasks or passed to subsequent GNN layers as a smaller coarsened graph. While computationally efficient, these approaches often ignore connections through removed nodes, leading to loss of global structural information.

\textbf{Cluster-based pooling methods} aim to preserve topological structure more explicitly by grouping nodes into clusters and coarsening the graph accordingly. For example, DiffPool \cite{ying2018hierarchical} uses learnable clustering matrices to coarsen each cluster to a single node as the representation of the corresponding community. MinCutPool \cite{bianchi2020spectral}, hinted by Spectral Clustering, minimizes minCut objective to cluster the nodes according to both the graph topology and node features. SEP-G \cite{wu2022structural} is a deterministic pooling method that leverages structural entropy minimization to determine cluster assignments. The advantage of Cluster-based methods is that it creates a concise representation of the original graph and significantly reduce the computation cost in proceeding GNN layers. 

\subsection{A unified view of hierarchical pooling models}
\label{traditional_unified_view}

In this subsection we provide a formal, unified mathematical formulation of cluster‐based graph pooling, laying the groundwork for our subsequent analysis and proposed enhancements.

A graph with \(N\) nodes can be represented as \(\mathcal{G} = \{ \mathcal{V}, \mathcal{E} \}\) with node set \(\mathcal{V}=\{v_i\}_{i \le N}\) and edge set \(\mathcal{E}\). Let \(A \in \{0, 1\}^{N \times N}\) be the adjacency matrix and \(X \in \mathbb{R}^{N \times d}\) be the node feature matrix. Given a set of graphs \(\{\mathcal{G}_i \}\) and corresponding labels \(\{y_i\}\), the task of graph classification is to take each \(\mathcal{G}_i\) as input and output the correct label \(y_i\).

Typically, a graph pooling layer divides the original graph into \(K\) clusters. The \(k\)th cluster of size \(N_k\) would have node list \(\Gamma^{(k)}\). Let the sampling operator \(C^{(k)} \in \mathbb{R}^{N \times N_k}\) and the assignment matrix \(S \in \mathbb{R}^{N \times K}\) be defined as
\begin{equation}
C^{(k)}[i, j] = 1 \text{ iff } \Gamma^{(k)}[j]=v_i,\quad
S[i, j] = 1 \text{ iff } v_i \in \Gamma^{(j)}
\end{equation}
In this case the node feature matrix of cluster \(k\) is the downsample of the original graph signal: 
\begin{equation}
X^{(k)} = C^{(k)T}X
\end{equation}
The pooling layer then coarsens each cluster to a single node. The adjacency matrix and graph signal of the coarsened graph are
\begin{equation}A_{coar} = S^TAS \in \mathbb{R}^{K \times K}\end{equation}
\begin{equation}X_{coar} = S^TX \in \mathbb{R}^{K \times d}\end{equation}
These graph features are fed to subsequent GNN layers and serve as a compact representation of the original graph, allowing progressive aggregation of graph structure while reducing computational complexity.

\begin{figure}
    \centering
    \includegraphics[width=0.90\linewidth]{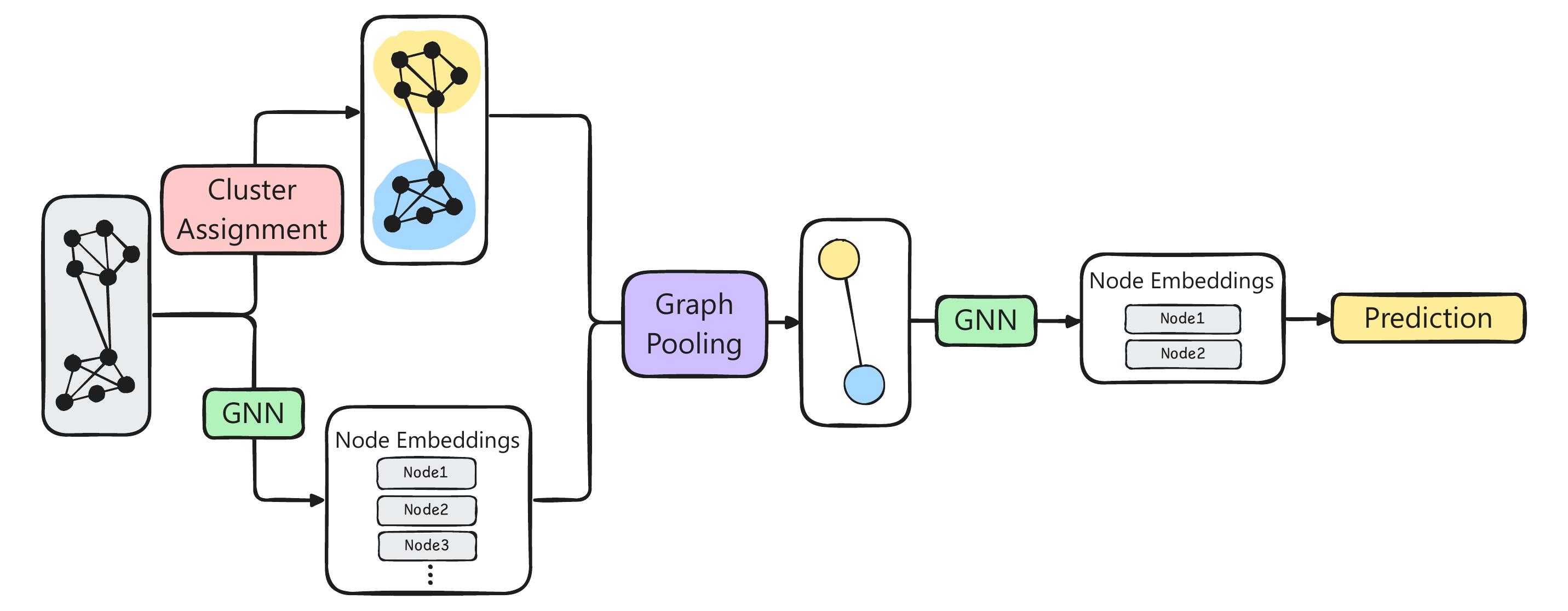}
\caption{General workflow of traditional graph pooling networks.}
    \label{fig:enter-label}
\end{figure}

\paragraph{Limitation of traditional pooling methods}

A critical limitation of existing pooling techniques lies in their treatment of inter-cluster relationships. Node-based pooling methods \cite{gao2019graph, lee2019self, ranjan2020asap, zhang2018end}, in which edges related to unimportant nodes are simply deleted, lacks sufficient preservation of graph connectivity, leading to a loss of structural information that is essential for modeling complex graph interactions. While cluster-based methods \cite{baek2021accurate, bianchi2020spectral, ma2019graph, ying2018hierarchical, yuan2020structpool} effectively summarize the internal structure of clusters, they often oversimplify the connectivity of the coarsened graph as \(A_{coar} = S^TAS\), which is an undirected representation that only counts the effective number of connections between clusters. This oversight can hinder the GNN's ability to learn meaningful representations for tasks where the interplay between clusters is significant.

\section{Methods}

\label{methods}

We introduce ICEPool, a novel graph coarsening framework that comprises two key components: (1). connection entropy matrix, a representation of coarsened graph's connectivity for distinguishing inter-cluster connections, and (2). SVDPool, an advanced cluster coarsening technique designed for coarsening clusters while capturing the undirected nature of inter-cluster connectivity.

\subsection{Preliminaries}

We first formalize the notation for describing intra- and inter-cluster connectivity, extending the framework introduced in Section \ref{traditional_unified_view}.

For a graph \(\mathcal{G} = (\mathcal{V}, \mathcal{E})\) with adjacency matrix \(A\) and cluster assignment matrix \(S \in \{0, 1\}^{N \times K}\), the \textit{intra-cluster adjacency matrix} is defined as:

\begin{equation}A_{int} = (SS^T) \odot A\end{equation}

where \(\odot\) denotes Hadamard product. This captures all edges between nodes within the same clusters.

The \textit{inter-cluster adjacency matrix} is
\begin{equation}A_{ext} = A - A_{int}\end{equation}
representing all edges connecting different clusters.

The \textit{inter-cluster connection matrix} of cluster \(a\) to cluster \(b\) is \(A_{a \rightarrow b} \in \mathbb{R}^{N_a \times N_b}\) such that
\begin{equation}A_{a \rightarrow b}[i, j] = 1 \text{ iff } A[\Gamma^{(a)}[i], \Gamma^{(b)}[j]] = 1\end{equation}
i.e. the \(i\)th node in cluster \(a\) is connected to the \(j\)th node in cluster \(b\).

\subsection{Connection Entropy-augmented Graph Attention}

To enhance the expressiveness of coarsened graph representations, we propose integrating a novel measure of inter-cluster structure into attention-based aggregation mechanisms. This section introduces the motivation and definition of connection entropy, followed by its integration with graph attention networks.

\subsubsection{Connection Entropy}

\begin{figure}
    \centering
    \includegraphics[width=0.70\linewidth]{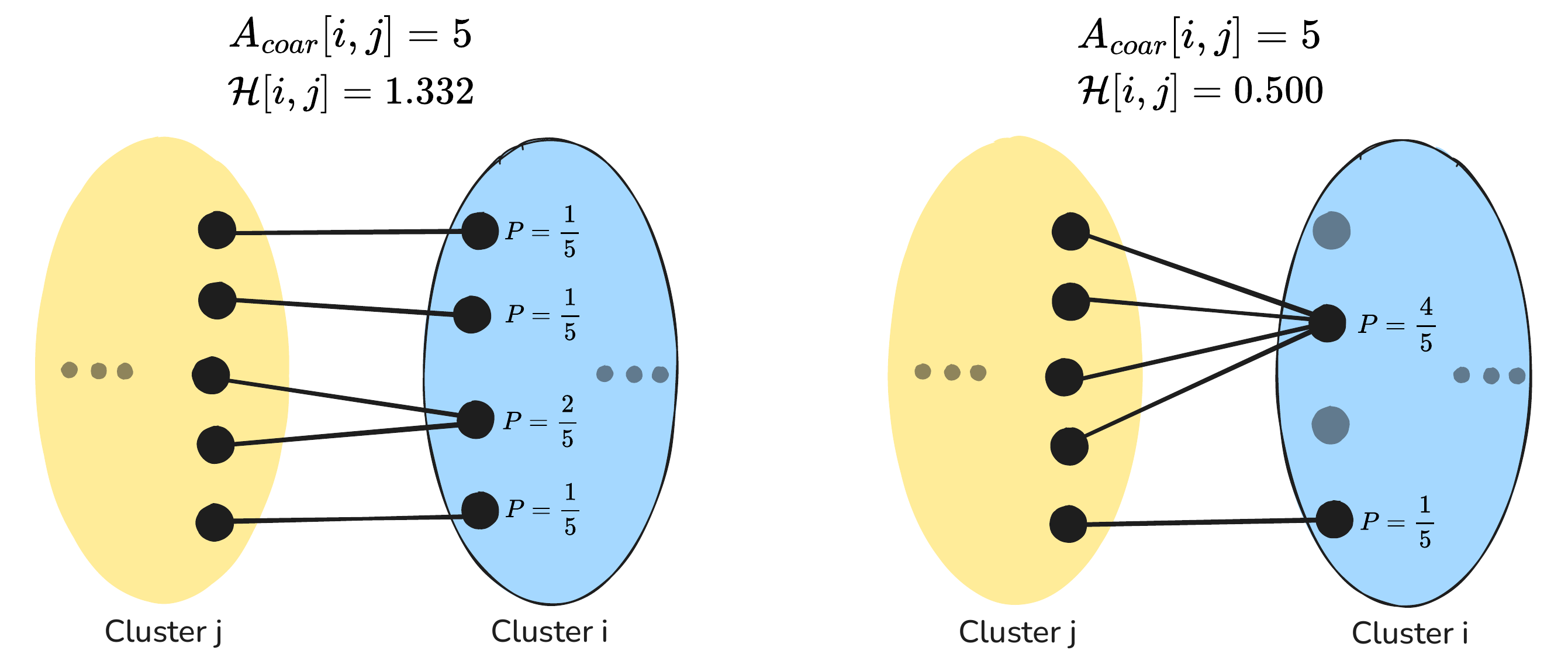}
\caption{Illustration of \textbf{Connection Entropy} \(\mathcal{H}\). Both graphs have the same number of inter-cluster edges between cluster \(i\) and cluster \(j\), resulting in identical coarsened edge weights \(A_{coar}[i, j] = 5\). However, the connection patterns differ: in the left example, edges are evenly distributed across nodes, while in the right, they are concentrated on a few nodes. The coarsened representation \(A_{coar}\) fails to capture this distinction due to its symmetry and reliance on edge count alone. In contrast, \(\mathcal{H}[i, j]\) reflects the distribution of connections, making it a more expressive measure of inter-cluster structure.}
    \label{fig:enter-label}
\end{figure}

\label{sec:connection entropy}

Representing the coarsened structure by \(A_{coar}\) often fails to capture the structural differences in clusters are connected. To address this limitation, we introduce a new measure called connection entropy, denoted by \(\mathcal{H}\), which quantifies the distribution of connections between clusters rather than just the count of edges. For two cluster \(i\) and \(j\), let \(P_{ij} \in \mathbb{R}^{N_i}\) be defined as:

\begin{equation}P_{ij}[n] = \frac{\sum_{k \in \Gamma^{(j)}} A_{i \rightarrow j}[n, k]}{\sum(A_{i \rightarrow j})}\end{equation}

where \(\sum(A)\) is the sum of all entries in \(A\). \(P_{ij}[n]\) can be interpreted as the probability of picking an edge with the \(n\)th node in cluster \(i\) at one end if we uniformly sample the edges between two clusters. The connection entropy matrix \(\mathcal{H} \in \mathbb{R}^{K \times K}\) is defined as:
\begin{equation}\mathcal{H}[i, j] = -P_{ij}^T \log P_{ij}\end{equation}

For a fixed number of edges between two clusters, \(\mathcal{H}\) assigns a lower value when connections are concentrated on a few nodes and a higher value when being evenly distributed across many nodes.

\subsubsection{Integration with Graph Attention}
To leverage our enriched edge features \(E = [A_{coar} \mathbin\Vert \mathcal{H} \mathbin\Vert \mathcal{H}^T]\) where \(\mathbin\Vert\) represents concatenation along the last dimension, we adopt two attention-based architectures:

\textbf{1. GAT with edge attribute:}
Graph Attention Network \cite{velivckovic2017graph} can incorporate edge features by concatenating them to node representations during attention computation:
\begin{equation}
    \alpha_{ij} = \text{softmax}\left( \text{LeakyReLU}\left( \mathbf{a}^T [\mathbf{W}\mathbf{h}_i \mathbin\Vert \mathbf{W}\mathbf{h}_j \mathbin\Vert \mathbf{W}_e E_{ij\cdot}] \right) \right)
\end{equation}

\textbf{2. Edge-Featured GAT (EGAT):} 
EGAT \cite{gong2019exploiting} explicitly integrates edge features through:
\begin{equation}
    \alpha = \text{DS}\left( \text{LeakyReLU}\left( \mathbf{a}^T [\mathbf{W}\mathbf{h}_i \mathbin\Vert \mathbf{W}\mathbf{h}_j] \right) \cdot E_{ijp} 
        \right)
\end{equation}
with Doubly Stochastic Normalization (DSN):
\begin{align}
    \tilde{E}_{ijp} &= \frac{E_{ijp}}{\sum_{k} E_{ikp}} \\
    \text{DS}(E)_{ijp} &= \sum_{k} \frac{\tilde{E}_{ikp} \tilde{E}_{jkp}}{\sum_{v} \tilde{E}_{vkp}}
\end{align}

In our architecture, we replace the conventional stack of multiple GNN layers after pooling operations by one GAT or EGAT layer.

\subsection{SVDPool}

While connection entropy effectively captures inter-cluster connectivity and provides a more informative representation, it solely focuses on the message passing in the coarsened graph and does not account for the pooling process. To further enhance the expressiveness of the coarsened graph, we introduce \textbf{SVDPool}, a method that utilizes singular value decomposition (SVD) to refine node representations in a way that adapts to the asymmetric structure of directed graphs.

\subsubsection{Motivation}
Spectral decomposition is widely recognized for its effectiveness in node clustering and graph classification tasks, particularly for undirected graphs. Notable examples are Spectral Clustering and EigenPool \cite{ma2019graph}. However, spectral analysis can't be readily generalized to connectivity between clusters. The connection entropy matrix introduced in the previous section highlights that asymmetric matrices provide a more accurate representation of coarsened graph's connectivity. This observation poses a fundamental challenge: The connection matrix \( A_{i \rightarrow j} \) between two clusters \( i \) and \( j \) is generally asymmetric, making eigendecomposition difficult to apply.  

\subsubsection{Directed pooling via SVDPool}

\begin{figure}
    \centering
    \includegraphics[width=1.0\linewidth]{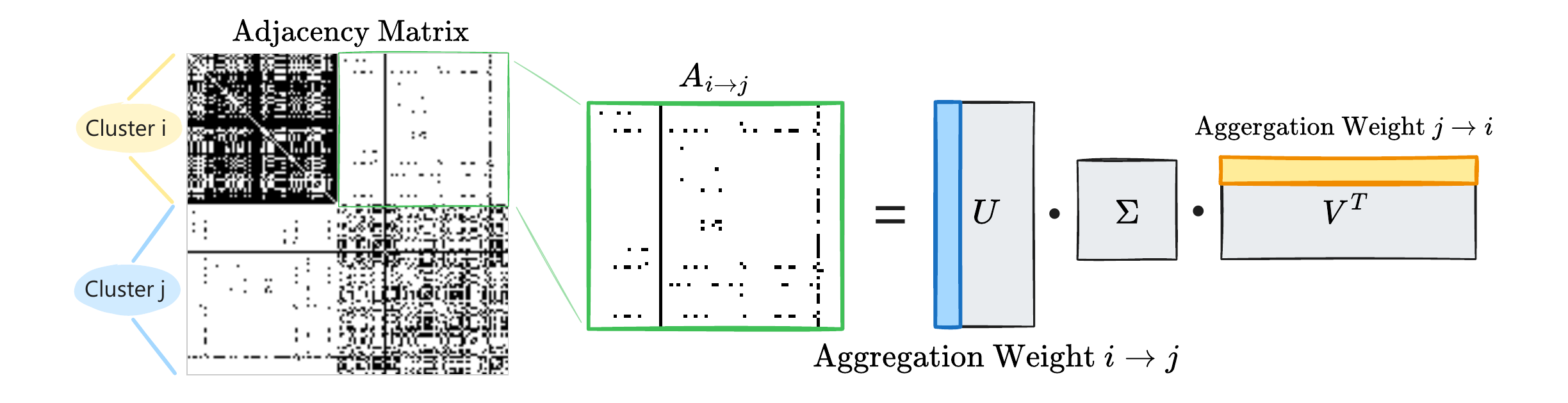}
\caption{Illustration of SVDPool. Node features are modulated by aggregation weights during message passing.}
    \label{fig:enter-label}
\end{figure}

Inspired by the success of \cite{wang2020spectral} and \cite{ma2019graph}, we introduce \textbf{SVDPool}, a novel approach that leverages singular value decomposition (SVD) to effectively model directed inter-cluster connections.

\textbf{SVDPool:} For connected clusters \(i\) and \(j\), let the SVD result for \(A_{i \rightarrow j}\) be
\begin{equation}A_{i \rightarrow j} = U_{i \rightarrow j}\Sigma_{i \rightarrow j} V_{i \rightarrow j}^T\end{equation}
For the \(l\)-th component, we first upsample the \(l\)-th column \(u_{i\rightarrow j}^l\), \(v_{i\rightarrow j}^l\) of \(U_{i \rightarrow j}\), \(V_{i \rightarrow j}\):
\begin{equation}\bar{u}_{i\rightarrow j}^l = C^{(i)}u_{i\rightarrow j}^l, \quad\bar{v}_{i\rightarrow j}^l = C^{(j)}v_{i\rightarrow j}^l\end{equation}
And the node aggregation is the weighted sum of nodes features:
\begin{equation}X_{i \rightarrow j}^l = \left(\bar{u}_{i \rightarrow j}^l\right)^TX, \quad X_{j \rightarrow i}^l = \left(\bar{v}_{i \rightarrow j}^l\right)^TX\end{equation}
Let \(\mathcal{U}^l \in \mathbb{R}^{N \times N \times N}\) such that:
\begin{equation}
\mathcal{U}_{ij}^l =
\begin{cases} 
\bar{u}_{i\rightarrow j}^l & \text{if } i\ne j\text{ and } j \in \mathrm{Neighborhood}(i), \\
0 & \text{otherwise }
\end{cases}
\end{equation}

To aggregate the whole neighborhood, we take the sum over all clusters:
\begin{equation}X_{\cdot \rightarrow j}^l = \sum_{i}^K\left(\mathcal{U}_{ij}^l\right)^TX\end{equation}
Finally, the graph signal after SVDPool is 
\begin{equation}Y^l = \Big\Vert_j X_{\cdot \rightarrow j}^l = \mathbf{U}^l X \quad\text{ where } \quad\mathbf{U}^l = \Bigg\Vert_j\left(\sum_{i}^K\left(\mathcal{U}_{ij}^{l}\right)^T\right)\end{equation}

To ensure scalability and robustness, several refinements are incorporated into SVDPool:
\begin{enumerate}

\item \textbf{Low-rank approximation.} Only the top \(H\) singular components are retained to reduce computational overhead while preserving the most informative directional patterns. If fewer than \(H\) components are available, the missing vectors are set to zero.

\item \textbf{Extended neighborhood perception.} To capture indirect or multi-hop interactions, we extend the adjacency matrices by replacing \(A_{i \rightarrow j}\) with \(A'_{i \rightarrow j}\), defined as:
\begin{equation}
A'_{i \rightarrow j}[m, n] = 1 \quad \text{iff} \quad (A^p + A^{p-1})_{\Gamma^{(i)}(m), \Gamma^{(j)}(n)} = 1
\end{equation}
Intuitively, \(p\) defines the radius of each cluster's neighborhood, and a value around $3$ is observed to be sufficient based on experiments.

\item \textbf{Singular value weighting.} To better reflect the contribution of each component, singular vectors are weighted by the square root of their corresponding singular values:
\begin{equation}
u_l' = \sqrt{\sigma_l} u_l, \quad v_l' = \sqrt{\sigma_l} v_l
\end{equation}

\item \textbf{Precomputable aggregation weight.} As SVD is performed on inter-cluster adjacency matrices, the process is suited for deterministic pooling frameworks where coarsening assignments remain fixed and can be precomputed.

\end{enumerate}

\subsubsection{Method Analysis}

We will show how SVD is related to inter-cluster connectivity and how SVDPool enhances model's understanding of the adjacency matrix.

\textbf{Proposition: SVDPool Perfect Reconstruction.} If \(N_{max}\) components are used, then inter-cluster connection matrix \(A_{ext}\) can be perfectly reconstructed.

Proof: 
\begin{align*}
\sum_{l=1}^{N_{\text{max}}} \sum_{i\ne j}^N \mathcal{U}_{i,j,:}^l (\mathcal{U}_{j,i,:}^l)^T 
&= \sum_{l=1}^{N_{\text{max}}} \sum_{i=1}^N \sum_{\substack{j \ne i}}^N \bar{u}_{i \rightarrow j}'^l \, \bar{v}_{i \rightarrow j}'^{lT} \\
&= \sum_{i=1}^N \sum_{\substack{j \ne i}}^N \sum_{l=1}^{N_{\text{max}}} \sigma^l \, C^{(i)} u_{i \rightarrow j}^l \, v_{i \rightarrow j}^{lT} (C^{(j)})^T \\
&= \sum_{i=1}^N \sum_{\substack{j \ne i}}^N C^{(i)} \left( \sum_{l=1}^{N_{\text{max}}} \sigma^l u_{i \rightarrow j}^l v_{i \rightarrow j}^{lT} \right) (C^{(j)})^T \\
&= \sum_{i=1}^N \sum_{\substack{j \ne i}}^N C^{(i)} A_{i \rightarrow j} (C^{(j)})^T \\
&= A_{\text{ext}}
\end{align*}

The ability to perfectly reconstruct inter-cluster connectivity highlights a fundamental advantage of SVDPool. By deploying SVDPool on an existing method that effectively preserves intra-cluster connectivity, the model provides a more comprehensive representation, ensuring that the coarsened graph retains both local and global connectivity patterns. 

\subsection{ICEPool Architecture}  
\label{subsec:ice_arch}  

\begin{figure}[t]  
    \centering  
    \includegraphics[width=1.0\linewidth]{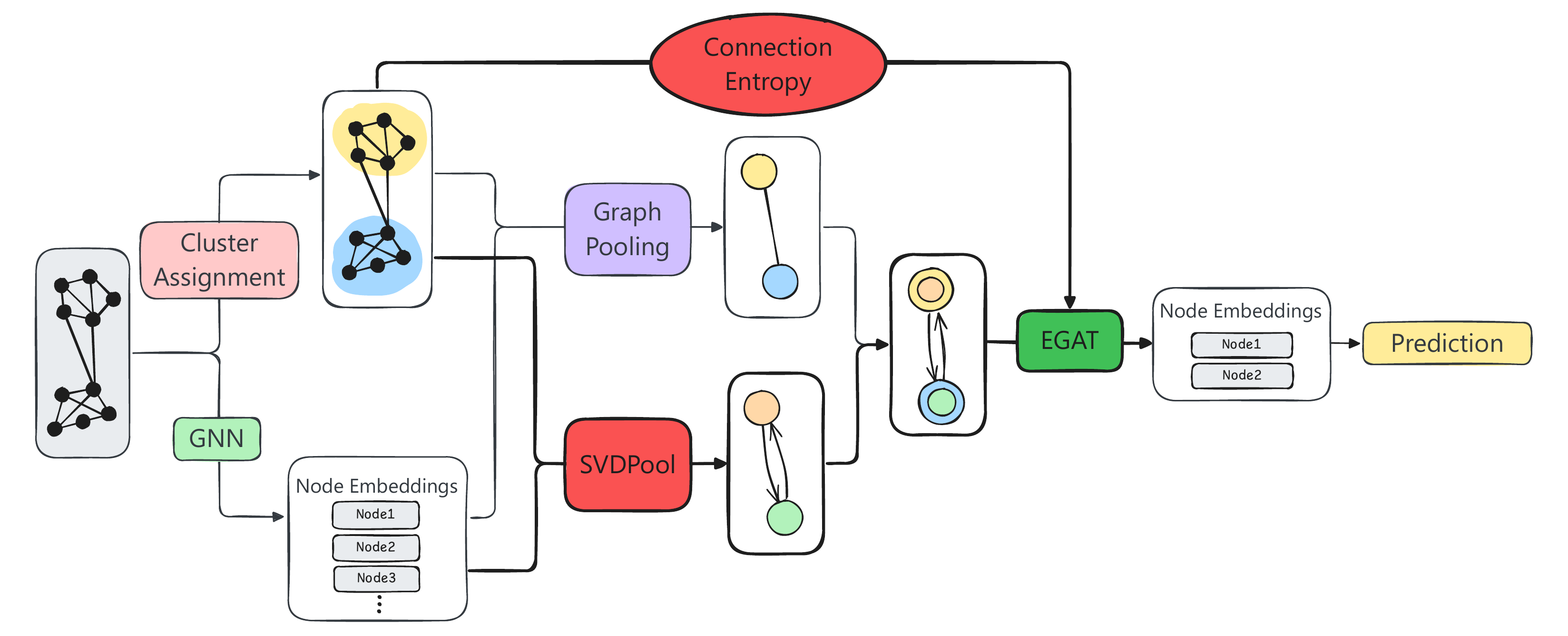}  
    \caption{ICEPool architecture. A SVDPool layer is added parallel to standard graph pooling layer, while the standard GNN layers that follows is replaced by a Connection Entropy-enhenced GAT layer.}  
    \label{fig:ice_arch}  
\end{figure}  

ICEPool extends traditional graph pooling frameworks through three stages (Figure \ref{fig:ice_arch}):  

\begin{enumerate} 
    \item \textbf{Preprocessing (optional)}: For deterministic clustering, connection entropy \(\mathcal{H}\) and SVDPool components are precomputed once from \(A\) and \(S\).  
    \item \textbf{Connection Entropy-enhanced GAT (CEGAT)}: Replaces standard GNN stacks after graph pooling layer with a single-head GAT/EGAT layer operating on edge features \(E = [A_{coar} \mathbin\Vert \mathcal{H} \mathbin\Vert \mathcal{H}^T]\).
    \item \textbf{SVDPool}: In addition to original graph pooling operation, SVDPool is applied to coarsen clusters while preserving asymmetric inter-cluster connectivity via spectral reconstruction of \(A_{ext}\).  
\end{enumerate}  

The architecture preserves the base model’s intra-cluster operations while augmenting inter-cluster processing. By separating intra/inter-cluster processing, ICEPool enhances structural fidelity without compromising the base model’s core functionality. The simple single-head attention layer and precomputation strategy guarantees the efficiency of the enhanced model.

\begin{table*}[ht]
\caption{Dataset Statistics}
\centering
\label{tab:dataset_stats}
\setlength{\tabcolsep}{3pt}
\renewcommand{\arraystretch}{1.0}
\begin{tabular}{lccccccc}
\toprule
 & \multicolumn{3}{c}{\textbf{Social Network}} & \multicolumn{4}{c}{\textbf{Bioinformatics}} \\
 \cmidrule(lr){2-4} \cmidrule(lr){5-8}
 & IMDB-BINARY & IMDB-MULTI & COLLAB & MUTAG & PROTEINS & D\&D & NCI1 \\
\midrule
\# Graphs & 1,000 & 1,500 & 5,000 & 188 & 1,113 & 1,178 & 4,110 \\
\# Classes & 2 & 3 & 3 & 2 & 2 & 2 & 2 \\
Avg. \# Nodes & 19.77 & 13.00 & 74.49 & 17.93 & 39.06 & 284.32 & 29.87 \\
Avg. \# Edges & 96.53 & 65.94 & 2457.78 & 19.79 & 72.82 & 715.66 & 32.30 \\

\midrule
GCN & 73.26 & 50.39 & 80.59 & 69.50 & 73.24 & 72.05 & 76.29 \\
GIN & 72.78 & 48.13 & 78.19 & 81.39 & 71.46 & 70.79 & \textbf{80.00}\\
Set2Set & 72.90 & 50.19 & 79.55 & 69.89 & 73.27 & 71.94 & 68.55\\
SortPool & 72.12 & 48.18 & 77.87 & 71.94 & 73.17 & 75.58 & 73.82 \\
SAGPool & 72.16 & 49.47 & 78.85 & 76.78 & 72.02 & 71.54 & 74.18 \\
StructPool & 72.06 & 50.23 & 77.27 & 79.50 & 75.16 & 78.45 & 78.64 \\
GMT & 73.48 & 50.66 & 80.74 & 83.44 & 75.09 & 78.72 & 76.35 \\
DiffPool & 73.14 & 51.31 & 78.68 & 79.22 & 73.03 & 77.56 & 62.32 \\
SAGPool & 72.55 & 50.23 & 78.03 & 73.67 & 71.56 & 74.72 & 67.45 \\
TopKPool & 71.58 & 48.59 & 77.58 & 67.61 & 70.48 & 73.63 & 67.02 \\
ASAP & 72.81 & 50.78 & 78.64 & 77.83 & 73.92 & 76.58 & 71.48 \\
MinCutPool & 72.65 & 51.04 & 80.87 & 79.17 & 74.72 & 78.22 & 74.25 \\
SEP-G & 74.12 & 51.53 & 81.28 & 85.56 & 76.42 & 77.98 & 78.35 \\
Cluster-GT & \textbf{75.10} & 52.13 & 80.43 & 87.11 & 76.48 & \textbf{79.15} & 78.64 \\
\midrule
\textbf{SEP-G ICE} & 74.90 & \textbf{52.47} & \textbf{81.76} & \textbf{87.22} & \textbf{77.21} & 78.44 & 79.54 \\
\bottomrule
\end{tabular}
\label{tab:graph-results-mean}
\end{table*}

\section{Experiments}
\label{experiments}

\subsection{Experiment Setup}

To demonstrate the versatility of ICEPool, we integrate it into three hierarchical pooling architectures: SEP-G, DiffPool, and DiffPool-det. All experiments are conducted using 10-fold cross-validation with Accuracy as evaluation metric. For ICE models, we retain the original hyperparameters of each baseline model and only fine-tune the parameters specific to ICEPool. Experiments are performed using NVIDIA A100 80GB PCIe and NVIDIA L40S GPUs.

We evaluated performance on seven graph classification benchmarks in TU datasets \cite{morris2020tudataset}. The following baseline pooling models are considered:
\begin{enumerate}
    \item Backbones: GCN and GIN \cite{kipf2016semi, wang2022powerful}.
    \item Global Pooling: Set2Set, SortPool, TopKPool, and SAGPool \cite{sutskever2014sequence, zhang2018end, gao2019graph, lee2019self}.
    \item Hierarchical Pooling: ASAP, StructPool, MinCutPool, GMT, and DiffPool \cite{ranjan2020asap, yuan2020structpool, bianchi2020spectral, baek2021accurate, ying2018hierarchical}.
    \item State of the Art Models: SEP-G and Cluster-GT \cite{wu2022structural, huang2024cluster}.
\end{enumerate}
For SEP-G, we apply a combination of CEGAT and SVDPool modules. We observed that placing SVDPool on the first layer yields the best results in most cases. This is reasonable since earlier layers contain more nodes and denser connectivity, where preserving inter-cluster structure is crucial.

\begin{table}[t]
\caption{Performance Comparison: SEP-G vs. SEP-G ICE}
\centering
\label{tab:sep-comparison-delta}
\setlength{\tabcolsep}{6pt}
\renewcommand{\arraystretch}{1.0}
\begin{tabular}{lccc}
\toprule
 & SEP-G & SEP-G ICE & Gain \\
\midrule
IMDB-BINARY & \(74.12 \pm 0.56\) & \(\textbf{74.90} \pm 0.85\) & +0.78 \\
IMDB-MULTI & \(51.53 \pm 0.65\)& \(\textbf{52.47} \pm 1.00\) & +0.94 \\
COLLAB & \(81.28 \pm 0.15\) & \(\textbf{81.76} \pm 0.81\) & +0.48 \\
MUTAG & \(85.56 \pm 1.09\)& \(\textbf{87.22} \pm 1.39\) & +1.66 \\
PROTEINS & \(76.42 \pm 0.39\) & \(\textbf{77.21} \pm 1.08\) & +0.79 \\
D\&D & \(77.98 \pm 0.57\)& \(\textbf{78.44} \pm 0.63\) & +0.46 \\
NCI1 & \(78.35 \pm 0.33\) & \(\textbf{79.54} \pm 0.85\) & +1.19 \\
\bottomrule
\end{tabular}
\end{table}

\begin{table}[!t]
\caption{Performance Comparison: DiffPool vs. DiffPool ICE}
\centering
\label{tab:sep-comparison-delta}
\setlength{\tabcolsep}{6pt}
\renewcommand{\arraystretch}{1.0}
\begin{tabular}{lcccc}
\toprule
 & DiffPool & DiffPool ICE (EGAT) & DiffPool ICE (GAT) & Gain \\
\midrule
MUTAG & \(79.29\pm 1.24\) & \(80.09\pm 0.87\) & \(\textbf{80.42}\pm 1.40\) & +1.13 \\
PROTEINS & \(73.24\pm 1.35\) & \(74.68\pm 1.56\) & \(\textbf{74.96}\pm 1.84\) & +1.72 \\
D\&D & \(77.95\pm 1.03\) & \(\textbf{78.21}\pm 1.18\) & \(76.24\pm 2.35\) & +0.26 \\
NCI1 & \(76.67\pm 1.88\) & \(77.18\pm 1.65\) & \(\textbf{78.44}\pm 1.17\) & +1.77 \\
NCI109 & \(\textbf{77.04}\pm 1.42\) & \(75.95\pm 1.91\) & \(75.80\pm 2.29\) & -1.09 \\
\bottomrule
\end{tabular}
\end{table}

For DiffPool, due to the high computational cost of applying SVD dynamically during pooling, we apply only the CEGAT module, using both vanilla GAT and EGAT. Results show consistent improvements across datasets, with GAT generally yielding greater boosts. An exception is D\&D, where ICE causes a performance drop. This suggests that the effectiveness may vary depending on the structural characteristics of the dataset.

\subsection{Ablation Study on DiffPool-det}

DiffPool-det is a variant of DiffPool that uses a single deterministic pooling layer \cite{dhillon2007weighted}, which makes it suitable for in-depth analysis. We conduct an ablation study by evaluating the base model and its combinations with CEGAT and SVDPool. 

\begin{table}[!htbp]
\caption{Performance Comparison: DiffPool-det vs. DiffPool-det ICE.}
\centering
\setlength{\tabcolsep}{4pt}
\renewcommand{\arraystretch}{1.0}
\begin{tabular}{lccccc}
\toprule
 & MUTAG & PROTEINS & D\&D & NCI1 & NCI109 \\
\midrule
DiffPool-det& \(78.47\pm 1.51\) & \(73.60\pm 1.07\) & \(76.15\pm 1.05\) & \(75.64\pm 1.09\) & \(74.43\pm 1.46\)\\
\midrule
+CEGAT (EGAT) & \(78.47\pm 1.35\) & \(73.87\pm 1.36\) & \(\textbf{76.32}\pm 1.78\) & \(77.60\pm 1.31\) & \(74.28\pm 1.32\)\\
+CEGAT (GAT) & \(79.95\pm 1.08\) & \(\textbf{75.32}\pm 2.61\) & \(72.39\pm 2.34\) & \(\textbf{77.98}\pm 1.34\) & \(75.19\pm 1.59\)\\
+SVDPool & \(79.80\pm 2.06\) & \(72.52\pm 2.41\) & \(76.16\pm 1.12\) & \(75.77\pm 1.58\) & \(74.32\pm 1.94\)\\
+Both (EGAT) & \(79.74\pm 1.09\) & \(73.24\pm 2.42\) & \(75.90\pm 1.15\) & \(76.67\pm 1.01\) & \(\textbf{75.75}\pm 1.13\)\\
+Both (GAT) & \(\textbf{80.37}\pm 1.29\) & \(74.68\pm 1.15\) & \(73.76\pm 3.91\) & \(77.20\pm 2.45\) & \(75.42\pm 1.90\)\\
\bottomrule
\end{tabular}
\label{tab:graph-results-mean}
\end{table}

The results suggest that the enhancements generally lead to better performance. It is worth noting that the combination of CEGAT and SVDPool does not consistently outperform the individual components, suggesting potential redundancy or interference when applied together.  

\section{Limitations}
\label{limitations}
Despite the effectiveness of the proposed methods, several limitations remain that suggest directions for future research:

\begin{enumerate}
    \item \textbf{Computational Overhead}: ICEPool introduces additional computational cost for calculating connection entropy and SVD components. Moreover, SVDPool imposes a substantial computational burden when applied to non-deterministic pooling methods.

    \item \textbf{Variance in Performance}: Empirical results indicate that ICEPool can increase the variance of accuracy across runs. Although average performance improves, the elevated variability may complicate reproducibility and make performance less stable.

    \item \textbf{Architectural Scope}: Further exploration of architectural variants---such as increasing the number of attention heads or integrating alternative aggregation strategies---may yield additional performance gains.  Future work should also investigate the compatibility of ICE framework with node clustering approaches.
\end{enumerate}

\section{Conclusion}
\label{sec:conclusion}

In this paper, we reveal the general limitations of current graph pooling methods and propose ICE framework to address the issues and retain the overlooked the inter-cluster structural information. In particular, we introduce SVDPool method to reconstruct the adjacency information when performing pooling operation and incorporate the connection entropy in the network to capture more precise inter-cluster structural information. Both theoretical analysis and experiment results prove the effectiveness of our methods. More importantly, our method can be applied to a wide range of pooling methods and provide an insight into reserving and leveraging the inter-cluster structural information.

\bibliographystyle{plainnat}
\bibliography{custom}

\end{document}